\documentclass{article}
\usepackage{ijcai23}

\usepackage{times}
\usepackage{soul}
\usepackage{xcolor}
\usepackage{url}
\usepackage[hidelinks]{hyperref}
\usepackage[utf8]{inputenc}
\usepackage[small]{caption}
\usepackage{graphicx}
\usepackage{amsmath,amssymb}
\usepackage{amsthm}
\usepackage{booktabs}
\usepackage{algorithm}
\usepackage{algorithmic}
\usepackage[switch]{lineno}
\usepackage{epsfig}
\usepackage{csquotes}

\usepackage{siunitx}

\DeclareMathOperator*{\argmin}{argmin}

\usepackage{latexsym}

\title{Reinforcement Learning-based Wavefront Sensorless Adaptive Optics Approaches for Satellite-to-Ground Laser Communication}

\author{
Payam Parvizi$^1$\and
Runnan Zou$^1$\and
Colin Bellinger$^{2,3}$\and
Ross Cheriton$^2$\And
Davide Spinello$^1$
\affiliations
$^1$Department of Mechanical Engineering, University of Ottawa, Ottawa, Ontario, Canada\\
$^2$National Research Council of Canada, Ottawa, Ontario, Canada\\
$^3$Faculty of Computer Science, Dalhousie University, Halifax, Nova Scotia, Canada
\\
\emails
\{pparv056, rzou043\}@uottawa.ca,
\{colin.bellinger, ross.cheriton\}@nrc-cnrc.gc.ca,
dspinell@uottawa.ca
}

\begin{document}

\maketitle

\begin{abstract}
Optical satellite-to-ground communication (OSGC) has the potential to improve access to fast and affordable Internet in remote regions. Atmospheric turbulence, however, distorts the optical beam, eroding the data rate potential when coupling into single-mode fibers. Traditional adaptive optics (AO) systems use a wavefront sensor to improve fiber coupling. This leads to higher system size, cost and complexity, consumes a fraction of the incident beam and introduces latency, making OSGC for internet service impractical. We propose the use of reinforcement learning (RL) to reduce the latency, size and cost of the system by up to $30-40\%$ by learning a control policy through interactions with a low-cost quadrant photodiode rather than a wavefront phase profiling camera. We develop and share an AO RL environment that provides a standardized platform to develop and evaluate RL based on the Strehl ratio, which is correlated to fiber-coupling performance. Our empirical analysis finds that Proximal Policy Optimization (PPO) outperforms Soft-Actor-Critic and Deep Deterministic Policy Gradient. PPO converges to within $86\%$ of the maximum reward obtained by an idealized Shack-Hartmann sensor after training of 250 episodes, indicating the potential of RL to enable efficient wavefront sensorless OSGC.

\end{abstract}

\section{Introduction}

\begin{figure}[t]
\centering
\includegraphics[width = 0.48\textwidth]{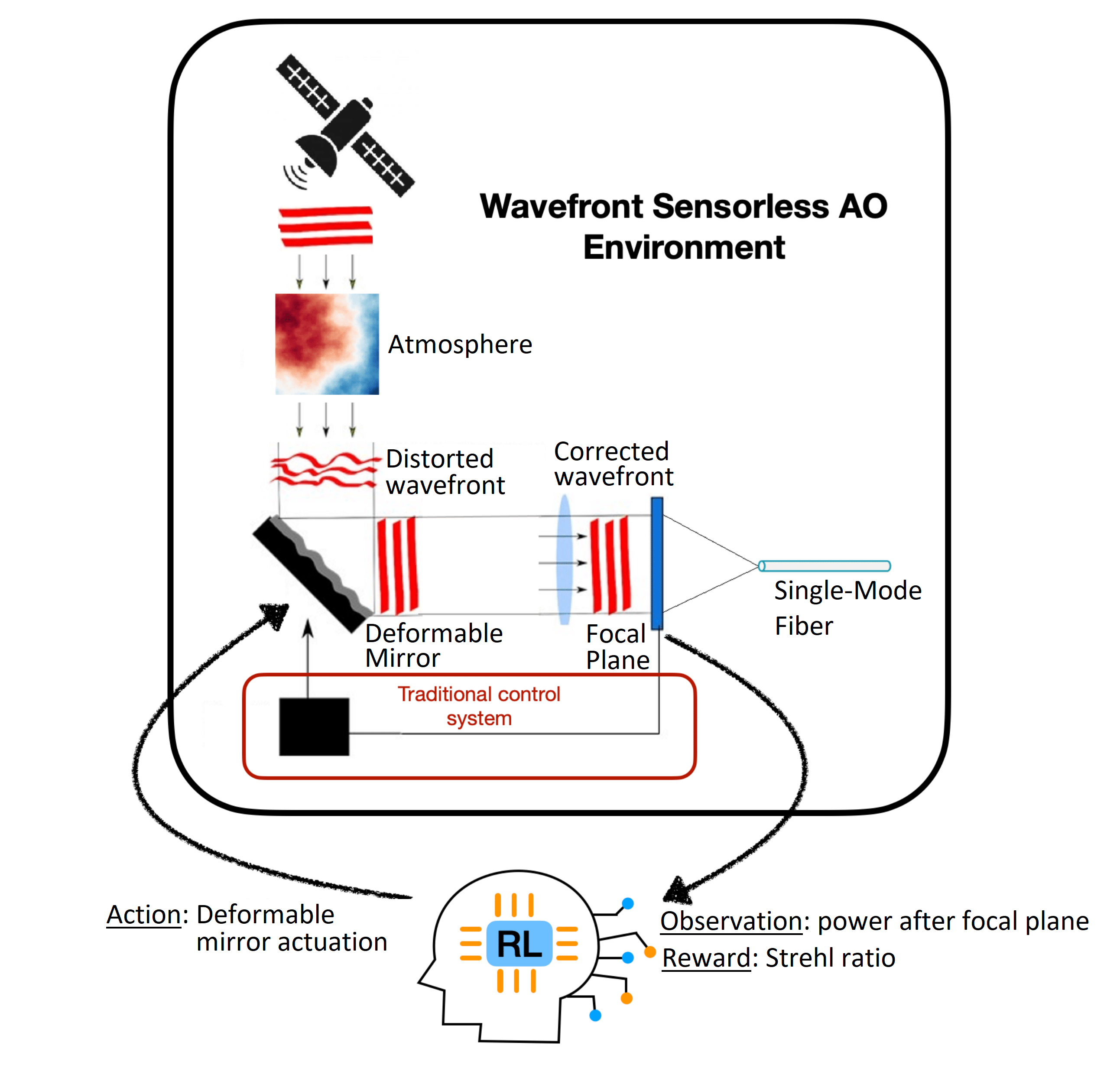}
\caption{Schematic of the RL environment of wavefront sensorless AO system}
\label{figs:rl_env}
\end{figure}

The internet has become an essential tool for education and commerce, yet it remains inaccessible or costly in many remote regions of the globe. Radio frequency satellite constellations are an existing solution that provide internet service; however, they experience a bottleneck in bandwidth due to the carrier wavelength of the link. Optical satellite-to-ground communication operates at near-infrared, thus, enabling a much faster data transfer rate. However, optical beams in optical communication can become distorted as they propagate through atmospheric turbulence, as illustrated in Figure \ref{figs:rl_env}, reducing the channel's potential bandwidth \cite{opt_com_space,atm_BER} 

Currently, most optical satellite communication ground stations have optical beam coupling into a single-mode fiber (SMF). This requires stronger optical beams from the satellite and a large telescope receiver, leading to high costs. Moreover, larger telescopes are more affected by atmospheric turbulence. This causes a diminishing return on a signal. The issue of atmospheric turbulence has been successfully mitigated in astronomy using adaptive optics (AO) methods \cite{wenhan2018ao,roddier1999adaptive,tyson2022principles}. The traditional AO method dynamically corrects distorted wavefronts in a feedback loop by using measurements from a wavefront sensor and applying them to adjust the distribution of actuators on the deformable mirror (DM), see Figure \ref{figs:AO_system}. 

Traditional AO systems are still costly and complex, with a significant portion of the cost arising from the wavefront sensor, especially when infrared beams are used for optical satellite-to-ground links. In addition, wavefront sensors in the infrared suffer from high read noise and require cooling and consume a fraction of the incident beam intensity. They have a limited dynamic range and introduce latency between measurements and the actuation of the DM. This can result in outdated wavefront measurements as the satellite rapidly moves across the sky. This introduces significant errors at the characteristic space-time scales. Recently, research has demonstrated the potential of using reinforcement learning (RL) to solve complex control problems in other domains of AO, such as astronomy \cite{ren2021alignment,tian2019dnn}. 

In this work, we propose to reduce cost and latency and improve accuracy using RL-based wavefront sensorless AO for optical satellite-to-ground links. This paper reports on the first phase of a 3-phase program that includes: \textit{a}) Researching and developing RL algorithms for wavefront sensorless AO in a simulated atmosphere, \textit{b}) Characterizing the RL algorithms through physical simulation, and \textit{c})  Deploying the RL model in a real-world AO system. Under the proposed setup, the RL agent learns to control the DM directly from the Strehl ratio of the power after the focal plane. The proposed RL environment is illustrated in Figure \ref{figs:rl_env}. 
 
In our empirical analysis, we compare Soft-Actor-Critic (SAC) \cite{haarnoja2018soft}, Proximal Policy Optimization (PPO) \cite{schulman2017proximal}, and Deep Deterministic Policy Gradient (DDPG) \cite{lillicrap2015continuous}, to an idealized traditional AO system with a Shack-Hartmann wavefront sensor on the developed RL environment. Our results suggest that RL can significantly improve coupling light into an SMF without a wavefront sensor under static environmental conditions.

To summarize, the contributions of this work are: 

\begin{itemize}
    \item The development of simulated wavefront sensorless AO RL environment for training and testing RL algorithms \footnote{\url{https://github.com/cbellinger27/adaptive_optics_gym}} 
    \item The first demonstration of the potential for RL in wavefront sensorless AO satellite-to-ground data links
\end{itemize}

The remainder of the paper is structured as follows: Section~\ref{sec:sdg} discusses the relevance of this work with the United Nations Sustainable Development Goals. Section~\ref{sec:trad_ao} includes background information on AO and related work on RL in the context of AO, and Section~\ref{sec:wsl_ao} details the RL environment developed as part of this work. The experimental setup and RL algorithms are described in Section~\ref{sec:method}, and Section~\ref{sec:results} presents the results. Finally, in Section~\ref{sec:discussion} we discuss the limitations of the RL environment and results, and Section~\ref{sec:conclusion} includes final remarks. 

\section{Relationship to the Sustainable Development Goals}\label{sec:sdg}

This research is relevant to SDG 9 \emph{(Building resilient infrastructure, promoting inclusive and sustainable industrialization and fostering innovation)}, as it aims to facilitate faster, more reliable, and lower-cost satellite-to-ground communications that can improve access to the internet in remote and rural regions. RL offers a promising solution for addressing this issue through improved light coupling with less latency at a reduced cost. We estimate that for every $1\%$ increase in light coupling, the cost of the system is reduced by $2\%$ \cite{belle2004telescopes,astrosysteme2022,planewaveobservatorysystems}.  In traditional AO, improving light coupling in a $40 \, \text{cm}$ telescope rather than using a $60 \, \text{cm}$ telescope results in savings of at least $50000 \, \text{USD}$ for the mount and telescope system \cite{planewaveobservatorysystems}, at least $30000 \, \text{USD}$ in the wavefront sensor, and at least $20000 \, \text{USD}$ in the dome enclosure. In contrast to traditional AO, adding an RL model to the system would only require a compact processor, such as a field-programmable gate array (FPGA), and a simpler detection unit, which would cost significantly less.


\section{Background}\label{sec:trad_ao}

\subsection{Adaptive Optics}

The objective of AO is to render and eliminate any phase distortions in the incoming light. The AO method was first proposed by \cite{babcock1953possibility} to improve astronomical images by correcting for atmospheric distortions with a deformable optical element controlled by a wavefront sensor. Since then, new techniques and results have been consistently published, primarily focusing on advancements in wavefront sensors and DMs \cite{bifano2011mems,corbett2007designing,nicolle2004improvement}.

\begin{figure}[t]
\centering
\includegraphics[width = 0.38\textwidth]{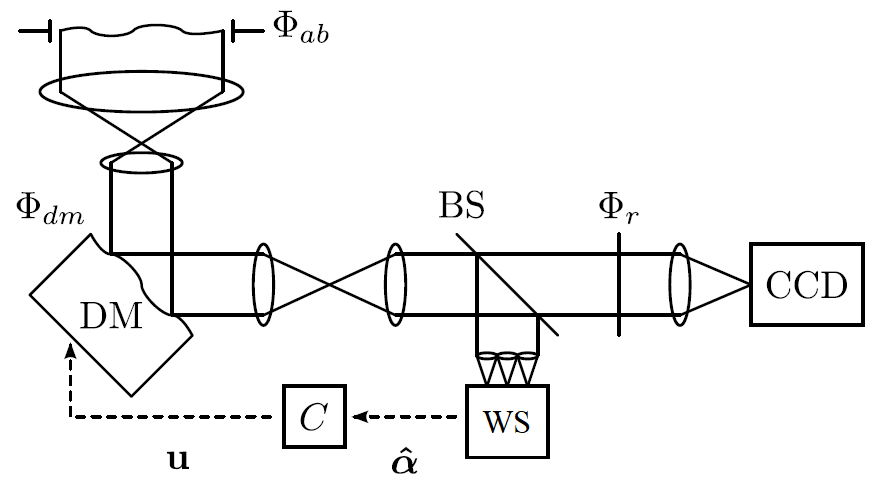}
\caption{Example of a traditional AO system for a telescope (recreated from \protect\cite{antonello2014optimisation}).}
\label{figs:AO_system}
\end{figure}

After propagating through the atmosphere to the aperture of a telescope, light is distorted by subtle changes in the temperature and pressure (and hence the index of refraction) of the air, which varies a function of time and space. The process of correcting the distortions with a traditional AO system is shown in Figure \ref{figs:AO_system}. At the entrance to the telescope, a phase distortion $\Phi_{ab}$ is present. Within the telescope, the light is projected onto a DM with $N_{a}$ actuators that create another phase aberration $\Phi_{dm}$. The phase aberration after the DM becomes $\Phi_r = \Phi_{ab} + \Phi_{dm}$. A beam splitter (BS) splits the light into two paths: one towards the exit pupil and the other towards the wavefront sensor (WS). On the path through the exit pupil, the image is created by focusing the light onto an exit pupil (detector) with the phase aberration of $\Phi_r$. The other path leads to the WS. This estimates the phase aberration $\Phi_r$ in the form of a set of Zernike coefficients $N_{\alpha}$ collected into a vector $\hat{\boldsymbol{\alpha}} \in \mathbb{R}^{N_{\alpha}}$. Finally, a controller (C) receives the coefficients and computes a vector $\mathbf{u} \in \mathbb{R}^{N_{a}}$, and creates a control signal of $N_a$ actuators of the DM. The main purpose of the controller is to minimize the phase aberration $\Phi_r$.

The objective of this work is to reduce cost and increase power through the use of RL. We hypothesize that RL can learn a control policy for the DM, thereby eliminating the need for the BS and WS, and reducing cost and increasing power.

\subsection{Satellite-To-Ground Communication}

The optical beam containing the communication signal can come from any one of a constellation of low earth orbit (LEO) satellites equipped with laser transmitters. Line-of-sight is a key required transmission between the telescope and a satellite. This is only maintained for a few minutes for a particular satellite, at which point the telescope must point to another satellite. 

The atmosphere has a characteristic turbulence timescale on the order of $\sim$$1 \, \text{ms}$, which varies with the satellite elevation angle. At the lowest elevations, the turbulence is the strongest, owing to the effective thickness of the atmosphere. The wavefront is also rapidly changing based on the change of optical path, which results in the turbulence profile appearing to translate across the aperture as the telescope is tracking the satellite. This means that the model must be able to determine the wavefront within $1 \, \text{ms}$ to maintain a high degree of correction as the wavefront changes while the satellite passes, and the adjustment of the DM must be made within tens of seconds. If an approximate solution can be found within a few milliseconds, the atmosphere can be considered to be in a quasi-static state and a static turbulence profile for the purposes of training can be considered valid.

\subsection{Reinforcement Learning (RL)}\label{sec:related_work}

RL is a machine learning technique that has been successfully applied to various continuous control tasks, including adaptive optics. In a wavefront sensor-based system, \cite{pou2022adaptive} proposed a multi-agent RL method for compensating for bandwidth error with a Shack-Hartmann sensor. \cite{nousiainen2021adaptive,nousiainen2022towards} applied model-based algorithms with a wavefront sensor in building an AO system to deal with the time delay error, and misregistration. \cite{pou2022model}, in another work, combined RL with a nonlinear reconstructor-based on neural networks in a U-net architecture for wavefront correction with a pyramid wavefront sensor. Within wavefront sensorless AO systems, DDPG and convolutional neural networks (CNN) were deployed to shift the performance of correction capacity, and speed for image sharpness \cite{keSelflearningControlWavefront2019a,hu2018build}. In these applications, CNNs extract features from images captured by a detector on the exit pupil. The DDPG then generates a control signal of DM based on CNN output. DDPG was also applied on a microscope by Durech in a wavefront system in which there is no atmospheric turbulence but lower speed turbulence from aqueous solutions and optical aberrations \cite{durech2021wavefront}. 

However, the existing implementations of RL on AO systems are insufficient for optical satellite communication due to their focus on optimizing image sharpness instead of optical data link reliability. Additionally, these implementations are optimized for different wavefront distortion conditions in environments such as microscopy and ophthalmology.

\section{Wavefront Sensorless Adaptive Optics RL Environment}\label{sec:wsl_ao}

The RL environment is implemented according to the standards of the Open AI Gym framework \cite{brockman2016openai}. The HCIPy: High Contrast Imaging for Python package \cite{por2018high} serves as the foundation of the RL environment. HCIPy offers a comprehensive set of libraries related to adaptive optics, including wavefront generation, atmospheric turbulence modeling, propagation simulation, fiber coupling, implementation of DMs and wavefront sensors. A simulated AO RL environment is a critical first step in the process of developing RL-based wavefront sensorless satellite-to-ground communication systems. It enables us and future researchers to assess and refine the RL to meet the strict requirements of this domain prior to costly evaluation in physical simulations and the real-world.

\begin{figure}[b]
\centering
\includegraphics[width = 0.47\textwidth]{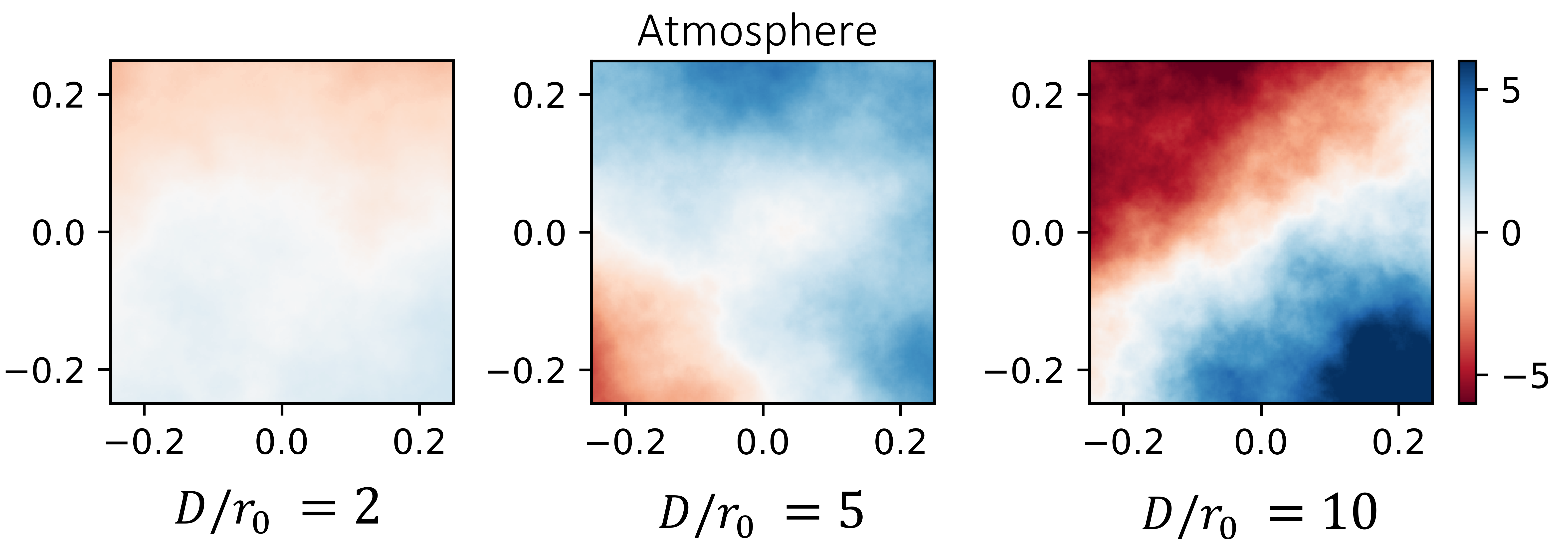}
\caption{Atmospheric turbulence conditions with respect to $D/r_0$}
\label{figs:layerss}
\end{figure}

The adaptive optics system simulated in this environment couples $1550 \, \text{nm}$ light into an SMF under different static turbulence conditions represented by the parameter $D/r_0$ (shown in Figure \ref{figs:layerss}), which is the ratio of the telescope's diameter $D$ to Fried's parameter $r_0$. This is a measure of the quality of optical transmission through the atmosphere. As Fried's parameter decreases, the transmission of light through the atmosphere becomes increasingly complex, resulting in more pronounced wavefront distortions. The $D$-value is fixed in this work to $0.5 \, \text{m}$, and the $r_0$-value is varied to assess the performance under different levels of atmospheric turbulence. The term ``static turbulence condition'' refers to the assumption that the turbulence in the atmosphere is stationary and the satellite is at a fixed position.

A graphical presentation of the RL environment is shown in Figure \ref{figs:rl_env}. The simulation environment updates in discrete time steps for practical purposes and consistency with the standard RL framework. As illustrated in the figure, at each time step $t$, the RL agent receives an observation of the system's current state ($o$) and a reward ($r$). The observation encodes the power after the focal plane in the AO system, and the reward is computed as the Strehl ratio of the power after the focal plane. Based on the agent's policy and the current observation, that agent selects the next action $\pi: o \rightarrow a$. The agent's actions control the DM in the AO system. If controlled optimally, the incoming optical beam becomes concentrated and centered on the SMF. 


\subsection{Episodic Environment}

Under real-world conditions, this RL problem can be characterized as a finite or infinite horizon problem. In the former, an episode lasts for the duration of the satellite's communication with the receiver. Here, we assess how efficiently an RL policy can transform the DM from its neutral position (flat) to a formation that focuses the beam on the SMF. Therefore, we define short (30, 50, and 100 time steps), fixed-length episodes starting from a neutral mirror.

\subsection{Action-Space}

\begin{figure}[t]
\centering
\includegraphics[width = 0.28\textwidth]{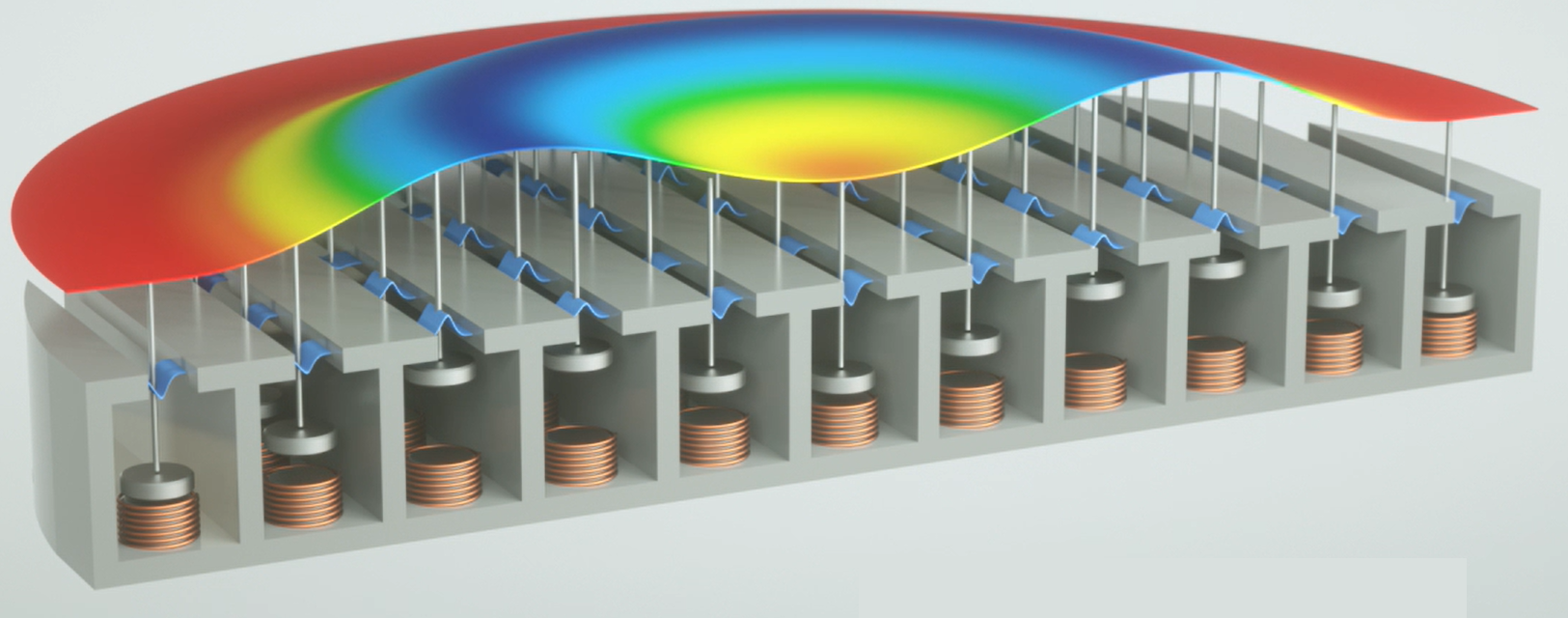}
\caption{Example of ALPAO DM surface and actuators. The image is presented by the ALPAO company \protect\cite{alpao}}
\label{figs:DM}
\end{figure}

In the context of AO, the RL agent's actions are movements of the actuators beneath the DM. The DM and actuators are illustrated in Figure~\ref{figs:DM}. The number of actuators determines the degree of freedom of the mirror's shape. The actuators are responsible for controlling the continuous reflective surface of the DM. The range of movement for these actuators is on the order of $\pm$$1 \, \mu \text{m}$, providing a high degree of precision in the mirror's surface shape and position.

The RL environment simulates a 64-actuator segmented DM, which has roughly 8 actuators across a linear dimension. Thus, the RL agent selects actions from a 64-dimensional continuous action-space where each actuator can be moved independently. This allows for smooth and precise control of the DM. The actuation has a limit corresponding to the maximum optical phase error that is possible under the atmospheric conditions used in training. The DM speed is assumed to be sufficiently fast to consider the atmosphere as quasi-static since most DMs are capable of correction speeds of up to a few kHz \cite{Pengwang2016}. Faster DMs are possible using smaller mirrors.

For a $50 \, \text{cm}$ telescope and this DM choice, the system can be expected to perform very well corrected for turbulence conditions of less than $r_0=6.25$.  We expect $r_0=0$ conditions to range from $5 \, \text{cm}$ to $15 \, \text{cm}$ for satellite elevation angles above $15^\text{o}$. 

\subsection{Observation-Space}

\begin{figure}[ht]
\centering
\includegraphics[width = 0.44\textwidth]{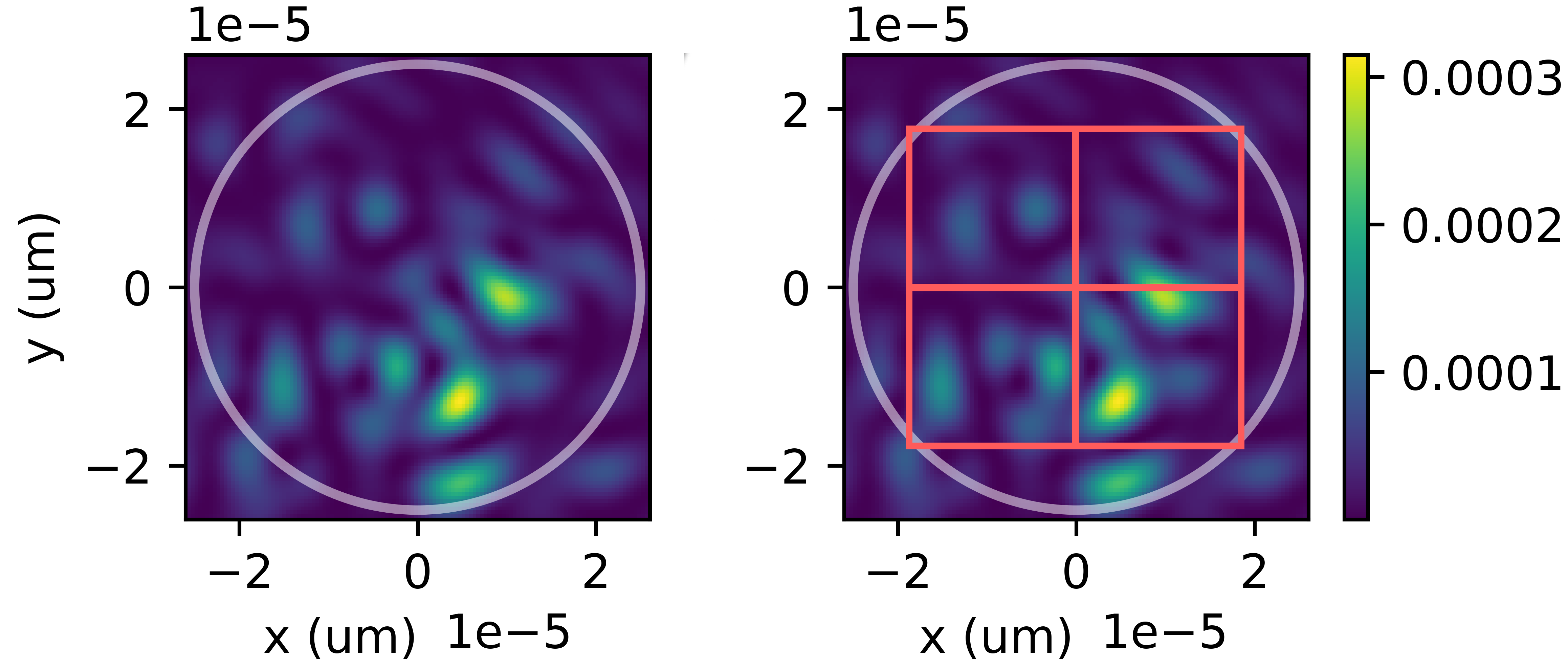}
\caption{Focal plane profile (left) continuous, (right) discretized}
\label{figs:beam_discrete}
\end{figure}

We utilize the power of the wavefront propagated through the focal plane to form the observation of the state of the environment. The focal plane is shown on the left in Figure~\ref{figs:beam_discrete}. We deal with observations rather than Markovian states because these can be directly and efficiently related to the light coupled into the fiber through a Strehl ratio calculation. Full state information would require access to information about the angle of the satellite and atmospheric conditions. 

For the observations, we discretize the focal plane into a sub-aperture array of $2\times 2$ pixels that can be realized with a fast and relatively low-cost quadrant photodetector, as shown on the right in Figure~\ref{figs:beam_discrete}. Using a low-pixel detector mitigates the use of slower and expensive read-out circuits used in infrared cameras, allows for more light per pixel for less noise, and improves the speed of the RL algorithm training.  

\subsection{Reward Function} \label{reward_func}

The reward function is calculated as the Strehl ratio of the optical system. It is defined as the ratio of the normalized peak intensity of the wavefront's point spread function (PSF) to the peak intensity of the ideal PSF without aberrations. A high Strehl ratio implies a high degree of wavefront correction, where a focused beam of light resembles an Airy disk and is approximately proportional to the amount of light that can be coupled into a fiber \cite{jovanovic2017aosmf}.  It is considered an approximation since a focused beam should resemble a Gaussian profile for optimal coupling into an optical fiber.  As proposed by Mahajan \cite{mahajan1983strehl}, the Strehl ratio of systems with a circular pupil is considered in terms of the variance of the phase aberration across the pupil. The expression is

\begin{equation} \label{eq1}
\mathrm{Strehl} = e^{-\sigma_{\Phi}^2}
\end{equation}

\noindent
where $\sigma_{\Phi}^2$ is the variance of the phase aberration. This quantity is chosen as the reward function since it is directly correlated to the amount of light coupled into an SMF \cite{jovanovic2017aosmf}.

\section{Experimental Methodology}\label{sec:method}

We quantify the RL performance as the mean and standard deviation of the Strehl ratio over 20 independent trials. Each RL algorithm has its hyperparameters tuned to the environment and the best-performing setup is compared to an idealized AO system with a Shack-Hartmann wavefront sensor. 

We implement and compare three families of RL algorithms: Soft Actor-Critic (SAC), Deep Deterministic Policy Gradient (DDPG), and Proximal Policy Optimization (PPO). These cover on- and off-policy learning, stochastic and deterministic policies and entropy-based methods, which are expected to have strengths and weaknesses in the context of wavefront sensorless adaptive optics. In particular, the off-policy algorithms, SAC and DDPG, generally have better sample efficiency than the on-policy algorithms. Alternatively, the on-policy method, PPO, is often more stable and easier to train. Given sufficient time, on-policy methods can provide good performance. Finally, the entropy regularization in SAC enables rich exploration, which is beneficial in high-dimensional action spaces. Each algorithm is discussed in more detail below. Table~\ref{tab:hyperparams} presents a comprehensive list of the hyperparameters selected for each method after tuning.

\begin{table}[ht]
    \centering
    \begin{tabular}{lrrr}
        \toprule
        Hyperparameter & SAC & DDPG & PPO \\
        \midrule
                  Buffer size                  & $128$      &  256        &  -          \\
                  Actor-${lr}$                 & $5e^{-4}$  &  $5e^{-5}$  &  $1e^{-2}$  \\
                  Critic-${lr}$                & $1e^{-2}$  &  $1e^{-2}$  &  $5e^{-6}$  \\
                  Actor-Hidden dim.            & $150$      &  250        &   150       \\
                  Critic-Hidden dim.           & $80$       &  65         &   50        \\
                  Clipping $\epsilon$          & -          &  -          &   0.35      \\
                  Temp. $\alpha_t$-${lr}$        & $1e^{-1}$  &  -          &   -         \\
                  Temp. $\alpha_t$-min limit     & $0.4$      &  -          &   -         \\
                  No episodes per iteration    & $1$        &  2          &   2         \\
                  No updates per iteration     & $20$       &  20         &   20        \\
                  Polyak ($\rho$)              & $0.99$     &  0.99       &   -         \\
                  Discount ($\gamma$)          & $0.95$     &  0.95       &   0.95      \\
                  Reward scaling               & No          & No           &   Mean-std  \\
                  learned $\alpha_t$           & Semi       &  -          &   -         \\                  
        \bottomrule
    \end{tabular}
    \caption{Hyperparameters and corresponding values}
    \label{tab:hyperparams}
\end{table}

\subsection{Soft Actor-Critic (SAC)}

SAC is an off-policy actor-critic algorithm based on a maximum entropy RL framework. It is particularly useful in complex and stochastic environments, such as wavefront sensorless AO systems. SAC uses a deep neural network to approximate the actor and critics. The actor component of the algorithm is tasked with maximizing the expected return while promoting exploration through random actions rather than becoming trapped in suboptimal policies. On the other hand, the critic component is responsible for estimating the Q-function of a given state-action pair. The Q-function provides feedback for improving the policy by adjusting the actions that the agent takes in each state to maximize the expected sum of future rewards \cite{haarnoja2018soft}.

SAC has shown promising results in various domains. However, one major drawback is its sensitivity to the choice of temperature ($\alpha_t$) and intuitively selected target entropy parameters. These parameters play a crucial role in the algorithm's performance, and their selection can significantly affect the outcome. To address this, \cite{haarnoja2018softb} proposed a method of automatic gradient-based temperature tuning by matching the expected entropy $\log \pi_{t}^{*}(\boldsymbol{a}_t |\boldsymbol{s}_t; \alpha_t) $ to a target entropy value $\Bar{\mathcal{H}}$ at time $t$
\begin{equation} \label{eq2}
\alpha_{t}^{*} = \argmin_{\alpha_t} \mathbb{E}_{\boldsymbol{a}_t \sim \pi_{t}^{*}} [\alpha_t \log \pi_{t}^{*}(\boldsymbol{a}_t |\boldsymbol{s}_t; \alpha_t) - \alpha_t \Bar{\mathcal{H}}],
\end{equation}
where the temperature $\alpha$ controls the stochasticity of the optimal policy. 

In our preliminary assessment, we evaluate SAC with a fixed temperature, learned temperature, and semi-learned temperature. The fixed temperature condition was optimized by setting a constant value of $\alpha_t = 0.4$, resulting in the best reward with the mean value of $52.63 \%$ and standard deviation of $14.64$ at the end of training. The learned temperature condition was optimized using the learning rate ${lr}_{\alpha_t} = 1e^{-1}$, with learning rates ranging from  \numrange[parse-numbers=false]{1e^{-6}}{5e^{-1}}. The semi-learned temperature condition was optimized using the same learning rate and a minimum alpha value of $0.4$, resulting in improved performance compared to the total range of learning rates and minimum $\alpha_t$ values between 0 and 1. The results show that the fixed and semi-learned learn at a faster rate than purely learned, and that semi-learned converges to the best policy. Thus, semi-learned is used in all subsequent experiments.

\begin{figure}[t]
\centering
\includegraphics[width = 0.43\textwidth]{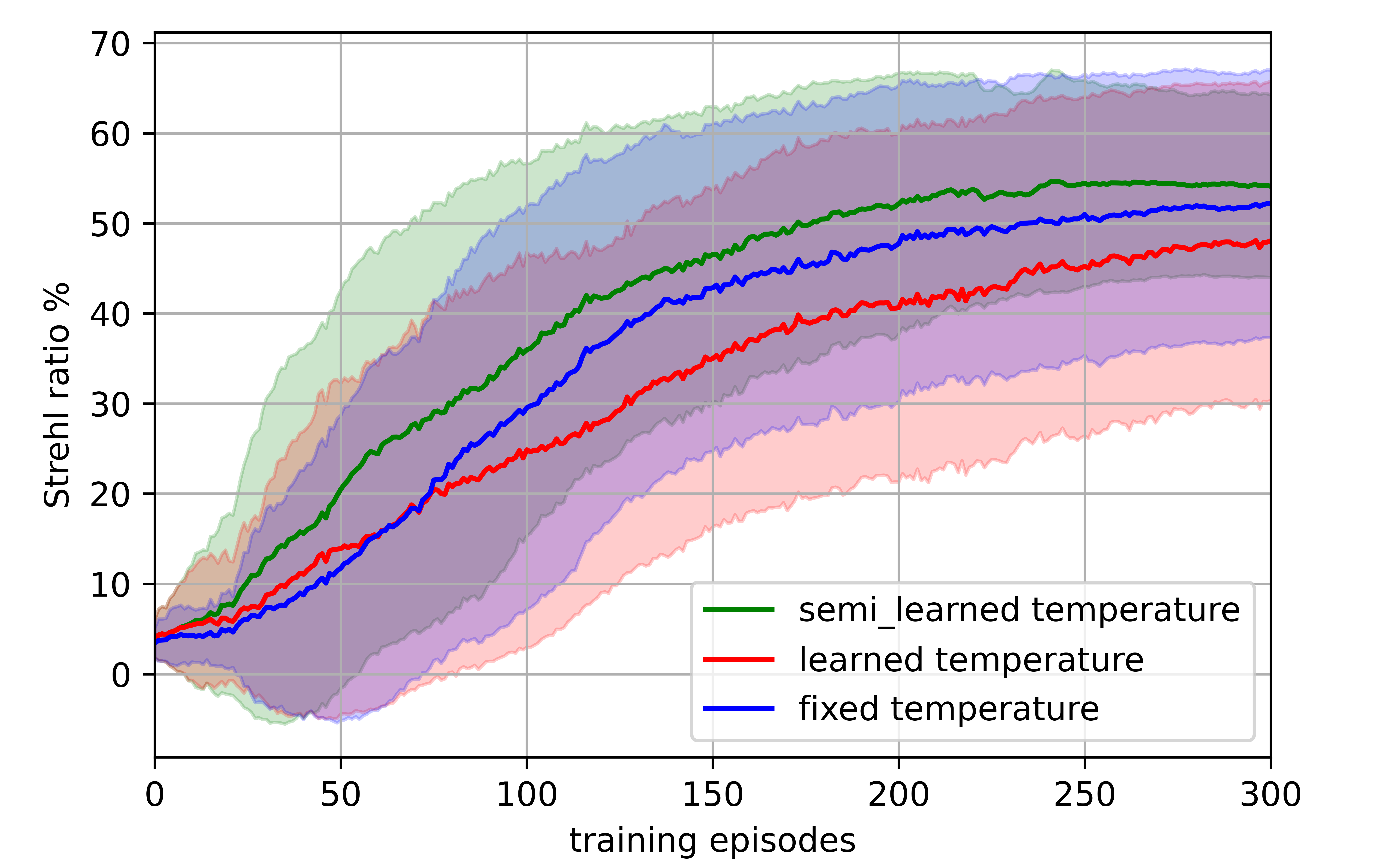}
\caption{Comparison of the selection of the temperature ($\alpha_t$) in SAC applied on 20 randomly selected static atmospheric turbulence of $D/r_0 = 5$}
\label{figs:comaprison_alpha}
\end{figure}

\subsection{Deep Deterministic Policy Gradient (DDPG)}

DDPG is an off-policy actor-critic algorithm for learning optimal control policies in continuous action spaces. It utilizes a deterministic policy, rather than a stochastic one, to map observations to actions. 


DDPG utilizes a deep neural network to approximate the value function and policy, allowing to handle high-dimensional state spaces. This approach, as previously demonstrated by \cite{lillicrap2015continuous}, can be effective in tackling complex environments. While DDPG has the advantage of ease of implementation, it can be sensitive to the choice of hyperparameters and can be prone to instability due to the choice of the reward function. 

In our preliminary assessment, we compared the effect of reward normalization with the mean-standard deviation (mean-std norm) and the min-max norm method versus no normalization on the learned policy. The normalization process scales the rewards across episodes. This has been shown to help the model identify actions that lead to higher rewards, thus accelerating the algorithm's convergence. In addition, as the model reaches convergence, the variance of the rewards tends to decrease, making it more challenging for the model to adjust itself. By normalizing the rewards, the model can more effectively recognize these rewards and continue to make adjustments.

\begin{figure}[ht]
\centering
\includegraphics[width = 0.43\textwidth]{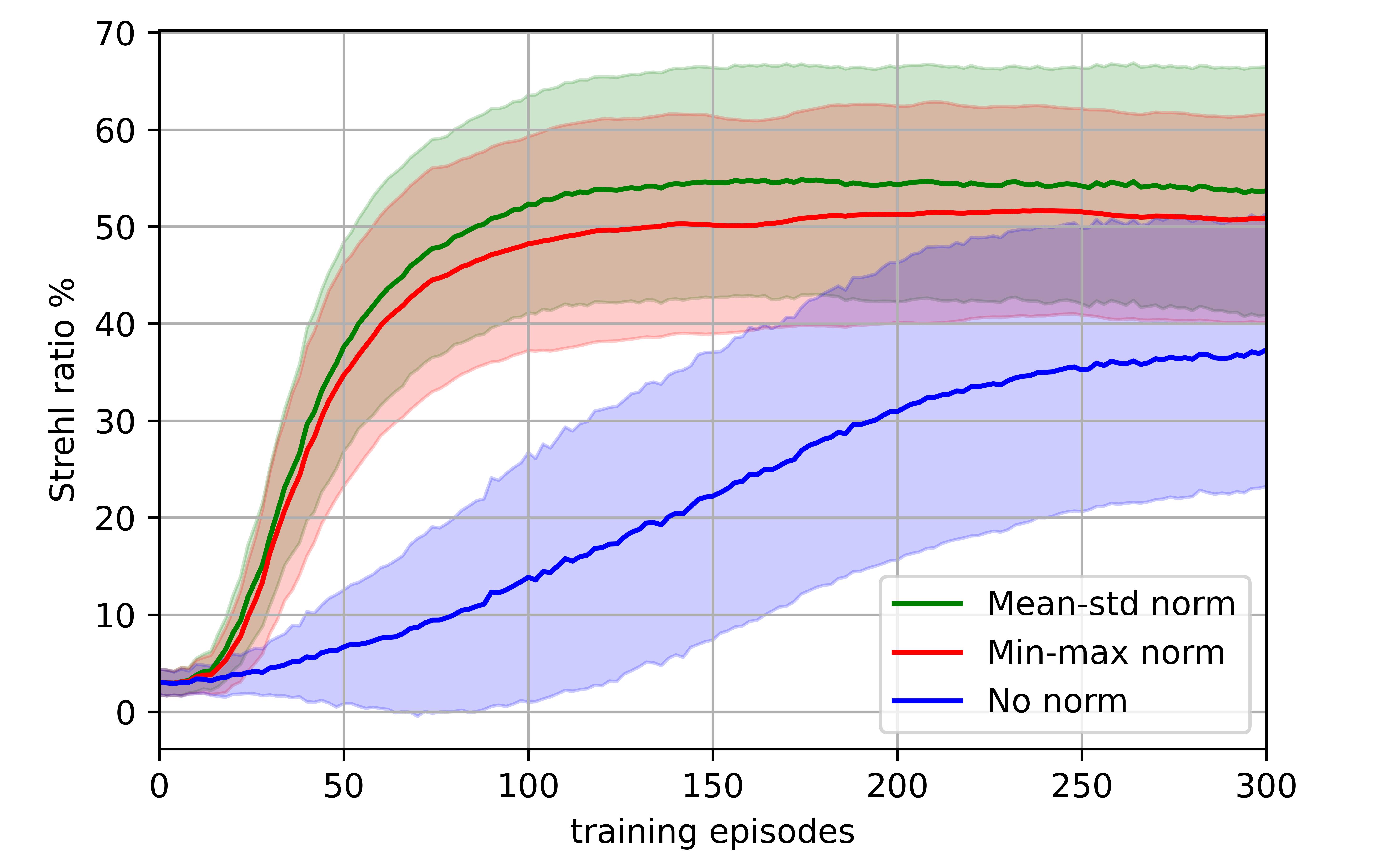}
\caption{Comparison of the selection of normalization technique in DDPG applied on 20 randomly selected static atmospheric turbulence of $D/r_0 = 5$.}
\label{figs:compare_norm_v1}
\end{figure}

Figure \ref{figs:compare_norm_v1} demonstrates that omitting reward scaling in this domain leads to slow convergence to a lower reward. Min-max norm and mean-std norm learn at similar rates, however, mean-std norm converges to a higher reward. Thus, mean-std normalization is employed for the subsequent experiments in our Deep Deterministic Policy Gradient (DDPG) approach. 

\subsection{Proximal Policy Optimization (PPO)}

PPO is an on-policy, policy gradient algorithm. It alternates between sampling data through interactions with the environment and optimizing a surrogate objective function via stochastic gradient ascent \cite{schulman2017proximal}. PPO utilizes a deep neural network to approximate the policy and value function. It uses a \enquote{clipped surrogate objective} to penalize significant changes to the policy. The policy regulation technique described in \cite{schulman2017proximal} limits the drastic changes of the policy during updates, but its effectiveness depends on the hyperparameter $\epsilon$, which controls the size of policy updates to prevent model collapse. Setting $\epsilon$ too small may result in slow convergence while setting it too large increases the risk of model collapse. 

Our preliminary analysis found that PPO is robust to $\epsilon \in [0.05, 0.4]$ in this environment. Settings within this range show very subtle differences in variance, convergence rate and convergence level. Generally, smaller $\epsilon$ values resulted in slightly slower convergence, whereas larger values converge faster but to marginally lower levels. Based on this analysis, all subsequent experiments have  $\epsilon=0.35$. 

\section{Results}\label{sec:results}




\subsection{Comparison of RL algorithms}

\begin{figure}[t]
\centering
\includegraphics[width = 0.45\textwidth]{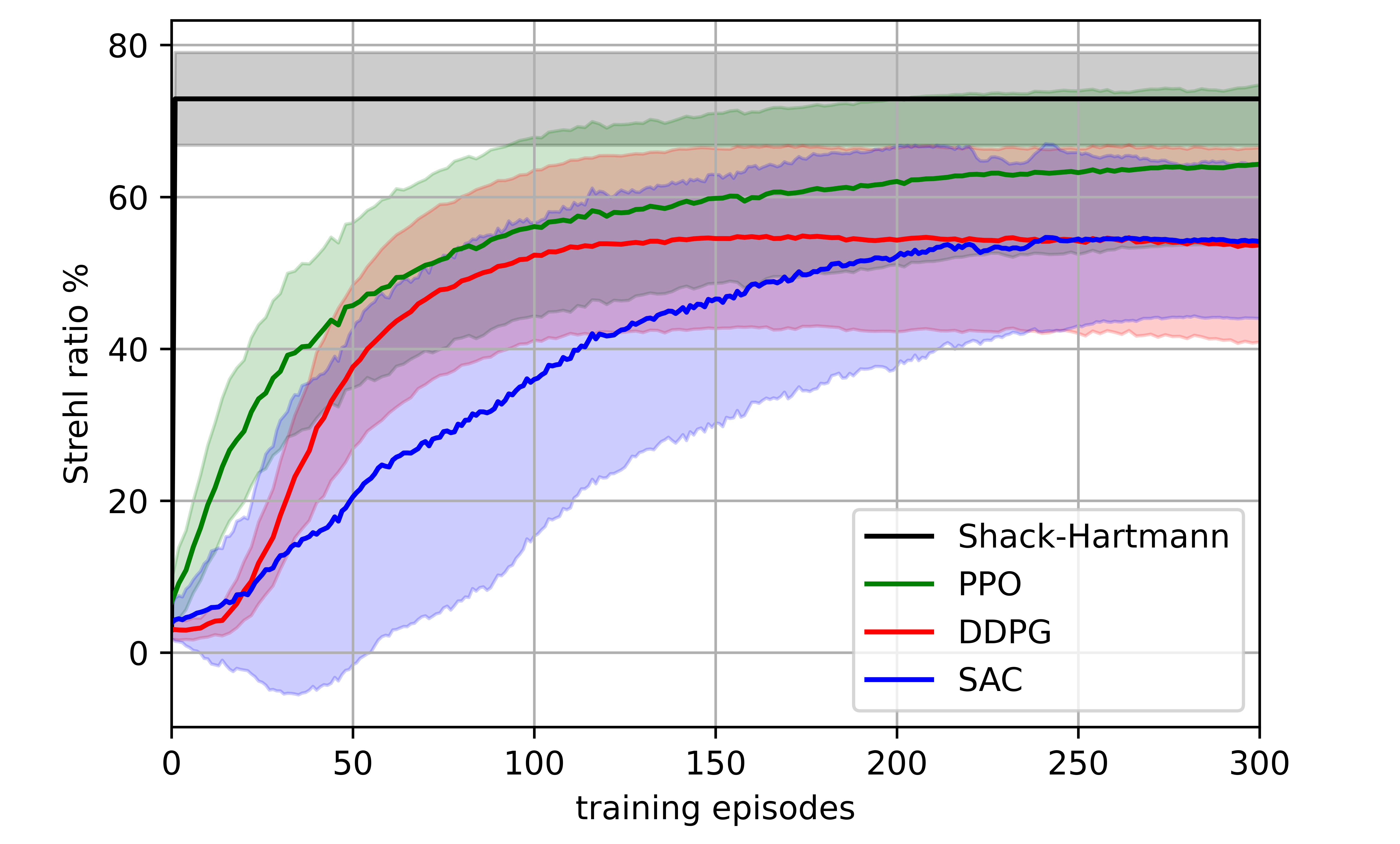}
\caption{Comparison of models applied on 20 randomly selected static atmospheric turbulence of $D/r_0 = 5$.}
\label{figs:comaprison_methods}
\end{figure}

We used the light coupling performance obtained from Shack-Hartmann wavefront sensor data as the reference for comparison with the refined RL algorithms outlined in Section \ref{sec:method}. This comparison was conducted under static turbulence condition of $D/r_0 = 5$. The results are presented in Figure \ref{figs:comaprison_methods}. Using the Shack-Hartmann wavefront sensor with 12 lenslets as benchmark allowed for a thorough evaluation of the effectiveness of the proposed RL algorithms in improving light coupling performance in the presence of atmospheric turbulence.

The results presented in Figure \ref{figs:comaprison_methods} indicate a better performance of the PPO algorithm in comparison to SAC and DDPG algorithms. Specifically, when considering the randomly selected static atmospheric turbulence of $D/r_0 = 5$, the PPO algorithm achieved a maximum reward of $67\%$, which is close to the maximum reward obtained by the Shack-Hartmann sensor, around $73\%$. While SAC and DDPG still demonstrated acceptable performance with a maximum reward of $53\%$ in the early training episodes, the PPO algorithm consistently demonstrated better results. The improved performance of PPO compared to SAC and DDPG can be attributed to its ability to sample from the action distribution, which helps avoid suboptimal local minima in high-dimensional state spaces. However, in SAC and DDPG, the actions that look nearly optimal can have an equal likelihood of being tried as those that look highly unoptimal.


\subsubsection{Performance with Turbulence severity}

\begin{figure}[tb]
\centering
\includegraphics[width = 0.43\textwidth]{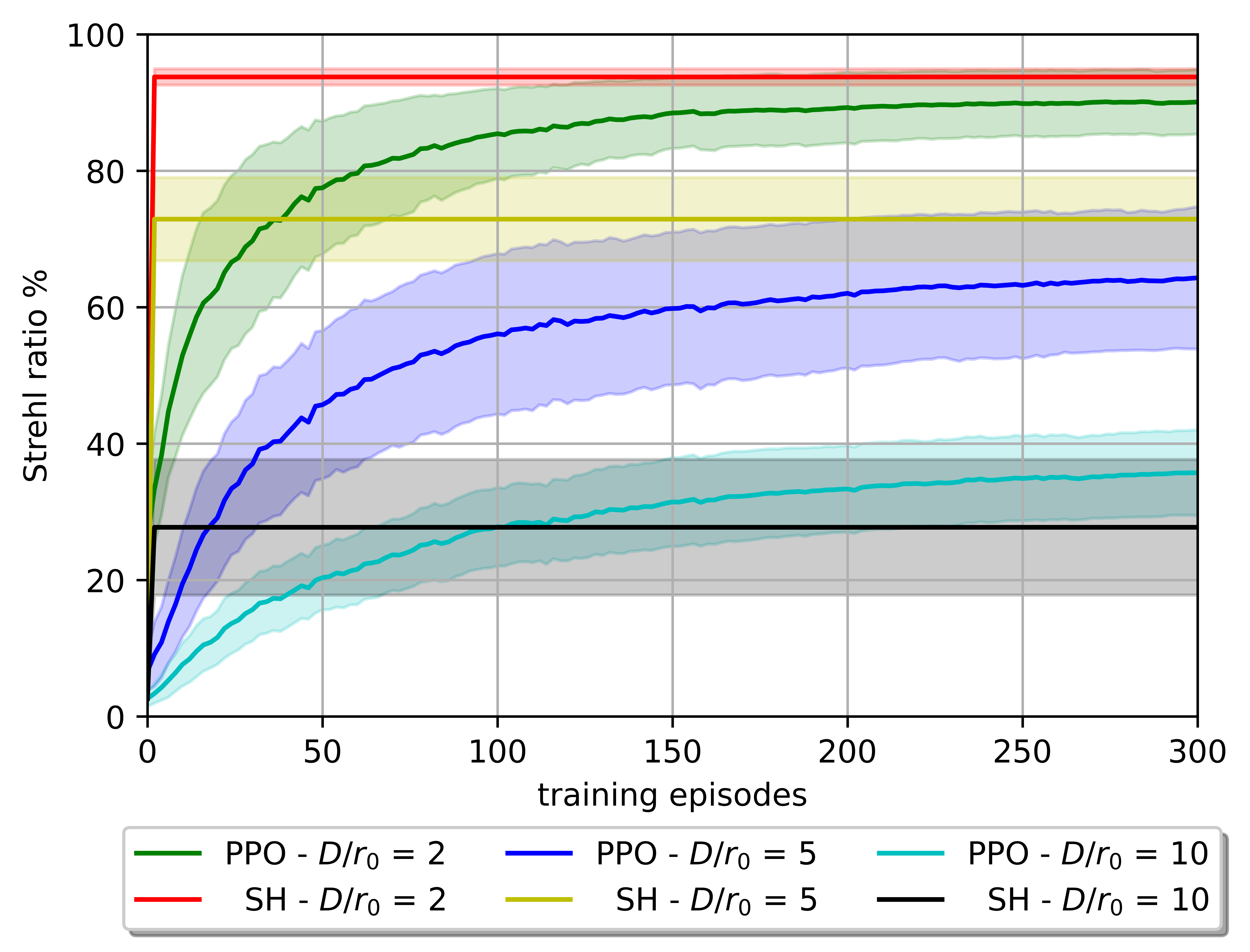}
\caption{Average reward of 20 randomly selected static atmospheres of different $D/r_0$ ratios with 64 actuators and 4 observers on PPO algorithm and Shack-Hartmann wavefront sensor}
\label{figs:ppo_SH_atm_compare_}
\end{figure}

Until this point, all experiments presented have been conducted under the turbulent condition of $D/r_0 = 5$. As previously stated, as the value of Fried's parameter decreases, there is a corresponding increase in the difficulty of transmitting light through the atmosphere. In this section, we used the PPO algorithm to assess its performance under mild and severe static turbulent conditions. The results are presented in Figure \ref{figs:ppo_SH_atm_compare_}.

As anticipated and illustrated in Figure \ref{figs:ppo_SH_atm_compare_}, a decrease in the value of Fried's parameter (or an increase in the ratio of $D/r_0$) results in a decline in the model's ability to attain a higher reward. If the agent performance cannot significantly improve beyond the uncorrected $2$ to $10\%$ Strehl ratio (depending on $D/r_0$), it can be considered impractical.

\begin{figure}[t]
\centering
\includegraphics[width = 0.40\textwidth]{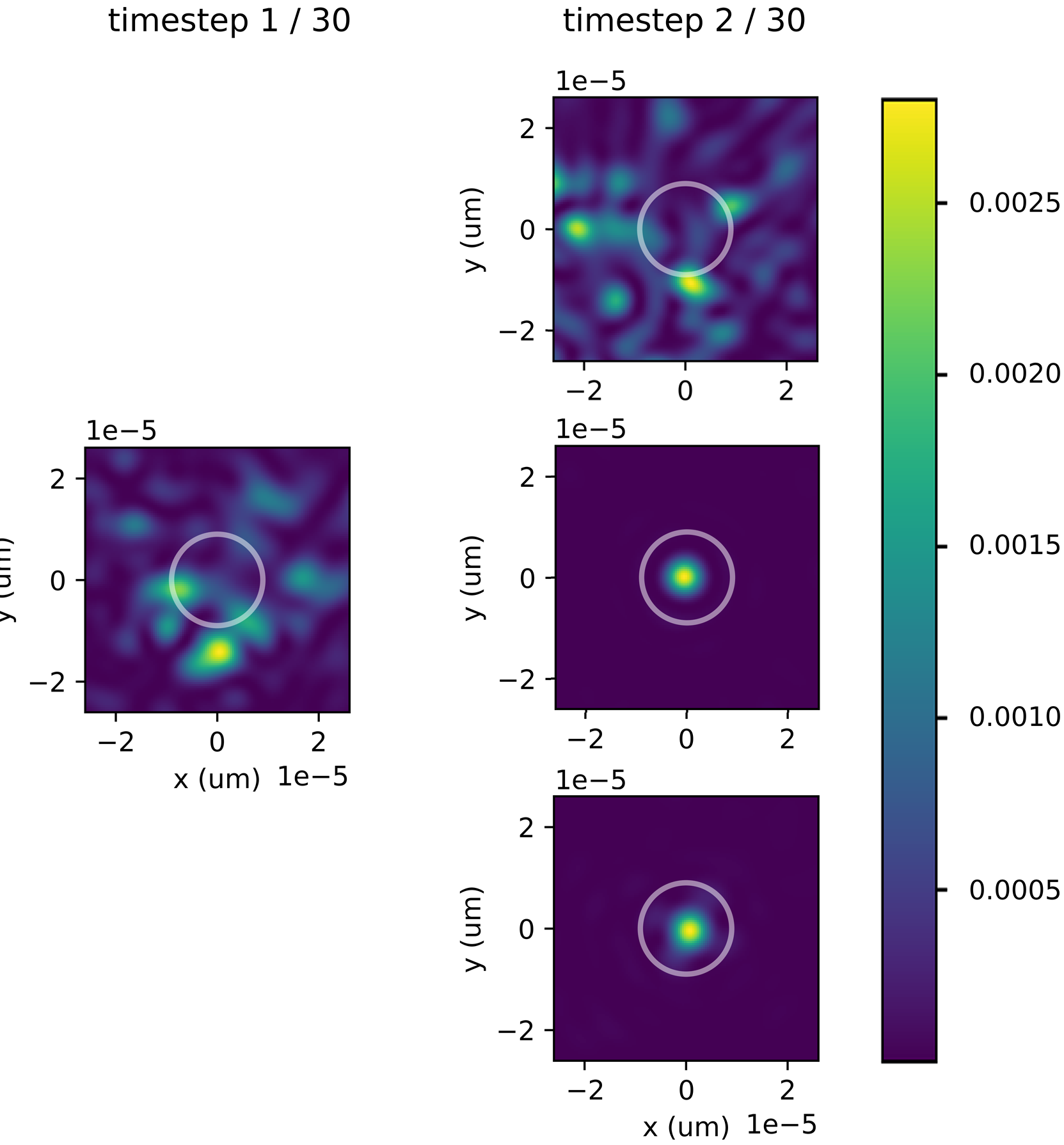}
\caption{Power distribution on a focal plane (left) before proposed AO, (right-up) after the implementation of random actions through the PPO algorithm in the initial episodes, (right-middle) after the utilization of the Shack-Hartmann wavefront sensor, (right-down) after the application of the PPO algorithm following a sequence of episodes.
}
\label{figs:before_after_ppo}
\end{figure}

The impact of using a PPO algorithm within an RL environment and using a Shack-Hartmann wavefront sensor on the power distribution of a wavefront in a static turbulence condition of $D/r_0 = 5$ have been analyzed and illustrated in Figure \ref{figs:before_after_ppo}. On the left of the figure, the power distribution at the focal plane at the start of each episode when the wavefront is reflected through a flat DM is illustrated. The right of the figure displays the results after (i) the implementation of random actions through the PPO algorithm in the initial episodes (upper right), (ii) the utilization of the Shack-Hartmann wavefront sensor (middle right), and (iii) the application of the PPO algorithm following a sequence of episodes (lower right).

The utilization of the Shack-Hartmann wavefront sensor, as illustrated in Figure \ref{figs:before_after_ppo} (right-middle), has resulted in a significant concentration of power at the center of the focal plane, with a Strehl ratio of approximately $70\%$. Similarly, the application of the PPO algorithm, as illustrated in Figure \ref{figs:before_after_ppo} (right-down), has also produced a significant concentration of power at the center of the focal plane; however, with a slightly lower Strehl ratio of approximately $60\%$.

\section{Discussion}\label{sec:discussion}


The results indicate that while it is possible for the PPO, SAC, and DDPG RL models to determine an accurate set of DM actions, the results generally underperform the output of the Shack-Hartmann wavefront sensor. PPO outperforms the SAC and DDPG models, converging on a Strehl ratio of over $60\%$ after hundreds of training episodes.  One of the main limitations of these results is the applicability to a dynamic atmosphere with limited deformable mirror speeds. The $3 \, \text{dB}$ bandwidth of the fastest mechanical deformable mirrors is limited to $<$$10 \, \text{kHz}$, which implies that less than 10 action-measurement loop iterations are required for the quasi-static turbulence condition to hold. For most DMs, this requirement cannot be met.  However, novel photonic chip-based phase corrector arrays are capable of speeds well in excess of $20 \, \text{kHz}$ \cite{momen_diab_spie2022} and can make use of a model requiring tens of action steps to converge.  

Despite this limitation, we expect such simplifications to be reasonable as an RL-based system may still outperform a Shack-Hartmann wavefront sensor with its corresponding and photon count requirements which are not considered in our comparisons. For existing and planned optical ground stations, adaptive optics may not be considered at all due to the cost and unreliability of wavefront sensing. Therefore, any improvement to the wavefront beyond an uncorrected case is still of value. Furthermore, the application of models trained on quasi-static turbulence profiles may be of value to RL environments with dynamically changing turbulence profiles for improvement to signal re-acquisition times. 

The choice in using a $2 \times 2$ quadrant photodetector as a source of feedback arises from its simplicity, cost and a high degree of correlation to improved SMF coupling. Using PPO, the Strehl ratio improves rapidly, but may be insufficient for achieving a reasonable degree of wavefront correction under high turbulence conditions because of the presence of a high-dimensional continuous action space in contrast to a low-dimensional observation space. 


\section{Conclusion}\label{sec:conclusion}

We present a reinforcement learning-based approach for wavefront sensorless Adaptive Optics in the context of optical satellite-to-ground communication. Specifically, we used off-policy algorithms like SAC and DDPG, as well as an on-policy algorithm, PPO, to achieve optimal coupling of $1550 \, \text{nm}$ light into a single-mode fiber under various static turbulence conditions. The results show that the PPO algorithm is particularly effective in achieving a high average Strehl ratio in a low number of training episodes. Furthermore, our approach eliminates the requirement of wavefront sensor measurements, thereby reducing the cost and latency of optical satellite-to-ground communication, making it a promising solution for fast and affordable internet access in remote and low-resources areas. Future work could include investigating the performance of Reinforcement Learning algorithms in various dynamic turbulence conditions at different times of the day.


\subsection*{Acknowledgements}

This research was supported by the National Science and Engineering Research Council (NSERC) of Canada through Discovery grant RGPIN-2022-03921, and by the National Research Council (NRC) of Canada through the AI4D grant AI4D-135-2.

\bibliographystyle{named}
\bibliography{main}




\end{document}